\documentclass[letterpaper, 10 pt, conference]{ieeeconf}  

\IEEEoverridecommandlockouts                              

\usepackage[numbers]{natbib}
\usepackage{multicol}
\usepackage[bookmarks=true]{hyperref}

\usepackage{amsmath}
\usepackage{amsfonts}       
\usepackage{mathtools}
\usepackage{nicefrac}       
\usepackage{times}
\usepackage[super]{nth}
\usepackage{siunitx}
\usepackage{bm}
\usepackage{xspace}
\usepackage{multirow}

\usepackage{booktabs}       
\usepackage{tabularx}
\usepackage{makecell}
\usepackage{arydshln}

\usepackage{adjustbox}
\usepackage{svg}
\usepackage{microtype}
\usepackage{graphicx}
\usepackage[font=small]{caption}
\usepackage{subcaption}
\usepackage{placeins}
\usepackage{xcolor}

\usepackage[ruled]{algorithm2e} 

\SetAlFnt{\small}
\SetAlCapFnt{\small}
\SetAlCapNameFnt{\small}
\SetAlCapHSkip{0pt}

\newcommand{\figref}[1]{Fig.~\ref{#1}}
\newcommand{\secref}[1]{Sec.~\ref{#1}}

\newcommand{\tabref}[1]{Tab.~\ref{#1}}

\newcommand{\methodname}{\text{MERLoc}\xspace}

\title{\LARGE \bf
\methodname: Robust Robot Localization in Dynamic Environments with Pose-Based Retrieval over 3D Gaussian Splatting References
}

\hypersetup{
  pdfauthor   = {Michael Baumgartner, Dario Mylonopoulos, Sammy Christen, Markus Gross, Moritz Bächer},
  pdftitle    = {MERLoc: Robust Robot Localization in Dynamic Environments with Pose-Based Retrieval over 3D Gaussian Splatting References},
  pdfsubject  = {Robot Localization},
  pdfkeywords = {Localization, 3D Gaussian Splatting, Dynamic Environments, Image Retrieval, Pose Estimation},
}

\author{Michael Baumgartner$^{1,2}$, Dario Mylonopoulos$^{2}$, Sammy Christen$^{2}$,
        Markus Gross$^{1,2}$, Moritz B\"{a}cher$^{2}$%
\thanks{$^{1}$M. Baumgartner and M. Gross are with ETH Zurich, Switzerland.}%
\thanks{$^{2}$M. Baumgartner, D. Mylonopoulos, S. Christen, M. Gross, and M. B\"{a}cher are with Disney Research, Switzerland.}%
}

\makeatletter
\let\@oldmaketitle\@maketitle
\renewcommand{\@maketitle}{\@oldmaketitle
  \vspace{0.5em}
  \begin{center}
    \includegraphics[width=\linewidth]{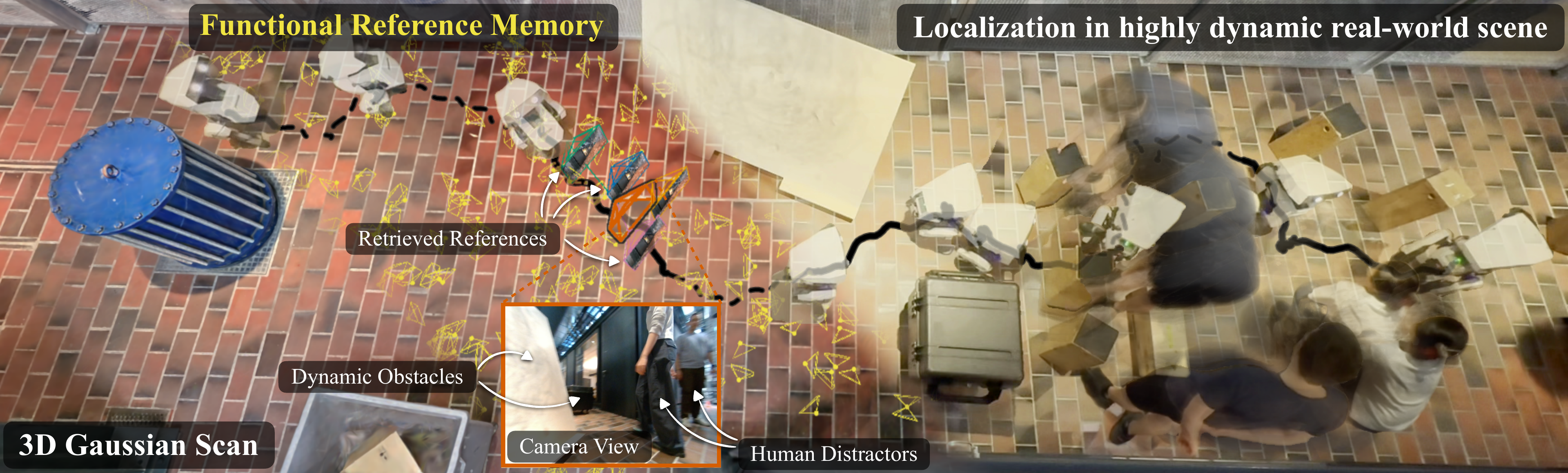}
    \setcounter{figure}{0}
    \captionof{figure}{\textbf{\methodname{} in a dynamic hallway.} 
    Overhead view: scanning the empty hallway (left), pre-processing the 3D Gaussian Splatting scene into a Functional Reference Memory (middle), and localizing in the runtime scene with moving people and displaced objects (right). Yellow frustums mark FRM reference poses. The black trace shows our onboard pose estimate, solid when localization succeeds and dotted when relying on odometry. For one timestep, we show the current pose estimate (vermillion frustum), the camera view, and the retrieved references in reddish purple, orange, blue, and bluish green.
    }
    \label{fig:teaser}
  \end{center}
  \vspace{-1em}}
\makeatother

\begin{document}

\maketitle
\thispagestyle{empty}
\pagestyle{empty}

\begin{abstract}

Robots operating in previously mapped spaces must localize accurately despite moving people and rearranged objects. Most localization pipelines match the query against a sparse set of captured reference images retrieved by appearance, which is brittle under scene change and repeating structures. We present \methodname, which turns a single metric 3D Gaussian Splatting scan into a dense \emph{Functional Reference Memory}. 
Reference views are sampled over the navigable space for the robot’s morphology and sensors at a target density and rendered offline. \emph{Pre-encoding} them for the chosen off-the-shelf localizer removes the reference encoder from inference.
A \emph{pose-based} retrieval strategy then selects references by spatial proximity and viewing direction rather than appearance, sidestepping visual ambiguities whenever a prior exists. Across one classical and four learned localizers in simulated and real dynamic scenes, pose-based retrieval matches or outperforms appearance retrieval in most cases, even when the latter is gated by the same prior. On real data, it lowers position RMSE by 15--60\,\% in nine of ten backend/split combinations, and the rendered memory outperforms real-image caches posed by Structure-from-Motion. 
The system runs onboard a Jetson Orin NX at \SI{5}{\hertz}, enabling a navigation policy to successfully traverse a hallway with people roaming and rearranging objects.

\end{abstract}

\section{Introduction}
\label{sec:intro}

Robots deployed in homes, hospitals, and warehouses operate in spaces that are easy to scan but rarely remain identical to those scans over time. People walk through them, objects appear, and furniture moves. While structural elements provide reliable localization cues, these dynamic changes introduce distractors and ambiguities that degrade localization pipelines built for static scenes. 

The dominant localization pipeline for previously mapped environments queries a sparse set of posed reference images~\cite{hloc}. References are posed offline by Structure-from-Motion (SfM)~\cite{colmap}, retrieved online via appearance descriptors~\cite{netvlad}, and passed to a relative localizer. 
Frozen at mapping time, the reference set is limited to the captured viewpoints and can neither be densified nor adapted to a robot’s sensor suite.
Furthermore, appearance-based retrieval is prone to \emph{visual aliasing}, where distinct places look alike, such as repeated corridors or identical doors. Aliasing and scene changes since mapping both lead retrieval to return confidently wrong references that no relative localizer can recover from~\cite{Lowry2016,Masone2021,sattler2018benchmark}.
Recent feed-forward localizers replace the geometric solver but inherit both limitations; even when trained on dynamic scenes~\cite{monst3r, vggt-omega, pi3}, most are, as we show, more sensitive than classical pipelines to aliased or changed references.

We propose a 3D Gaussian Splatting-based \textbf{M}emory \textbf{E}ncoding and \textbf{R}etrieval \textbf{Loc}alization (\methodname{}) framework, which replaces the rigid, captured reference set with a controllable, backend-adaptive, rendering-based memory. Given a 3D Gaussian Splatting (3DGS)~\cite{gross3dgs, kerbl3dgs} scan of the environment, we sample reference poses over the navigable space at a specified density, height range, and orientation profile tailored to the robot's sensor suite. We then render these views and encode them offline in a representation customized for the relative localizer. This forms a \emph{Functional Reference Memory} (FRM): its content is determined by the target backend and robot morphology rather than by the scanner that captures it, e.g., depth-lifted keypoints for classical pipelines~\cite{hloc} or encoded visual tokens for latent-conditioned architectures~\cite{monst3r,vggt-omega,pi3,reloc3r}. Because of this customized encoding, the relative localizer runs entirely off-the-shelf, without per-scene training or adaptation, and without encoding any reference at runtime, which removes one encoder pass per retrieved reference from the online path. Furthermore, the FRM density becomes a tunable design parameter that trades accuracy against storage without increasing online inference cost.

To select references, we index the dense FRM by pose. Given a pose prior, typically the previous estimate or odometry, we gate candidates within a radius and rank them by viewing-direction alignment and proximity, falling back to appearance-based matching only for initialization or recovery. Unlike SLAM, which routinely gates retrieval with a propagated pose~\cite{mcl,orbslam2}, localization pipelines still retrieve by appearance alone. 
Neither the dense FRM nor pose-based retrieval suffices alone: pose-based retrieval starves on a sparse reference set, and appearance retrieval still aliases on a dense one. Together, they make classical and learned relative localizers metric, more accurate, and more robust to scene change without touching backend weights.

We claim that (i) \methodname turns one metric 3DGS scan into a \emph{Functional Reference Memory} whose density, height band, and encoding are chosen per robot and per localizer, so that classical and feed-forward backends run off-the-shelf without encoding references at runtime (\secref{sec:system}); (ii) on real data, this rendered memory outperforms real-image caches of the same scene posed by SfM, even at a lower pose density for the classical backend (\secref{sec:exp_mocap}); (iii) \emph{pose-based} context retrieval over this memory lowers position RMSE by 15--60\,\% relative to appearance retrieval for nine of ten real-world backend/split combinations and, under simulated scene change and human activity, matches or exceeds prior-gated appearance retrieval while clearly exceeding plain appearance retrieval for four of five backends (Secs.~\ref{sec:exp_arena}--\ref{sec:exp_mocap});
(iv) the system runs at \SI{5}{\hertz} on a Jetson Orin NX, enabling a navigation policy to reliably traverse a hallway while people move around the robot and displace objects (\secref{sec:exp_nav}).

\section{Related Work}
\label{sec:related_work}

\subsection{Visual Localization and Aliasing}
\label{sec:rw_rel_loc}

The main approaches driving visual localization research are scene coordinate regression (SCR), absolute pose regression (APR), and retrieval-based localization with classical or learned relative pose estimation.

\subsubsection{Scene Coordinate and Absolute Pose Regression} SCR trains a network to predict per-pixel 3D scene coordinates~\cite{dsac-star,ace}; APR regresses the pose directly~\cite{posenet}. Both bake the scene into network weights and degrade under structural change~\cite{rio10}; APR moreover behaves like pose interpolation~\cite{sattler2019apr}. \methodname trains nothing per scene and runs off-the-shelf, scene-agnostic backends over an explicit reference memory.

\subsubsection{Retrieval-Based Localization}

Classical structure-based pipelines and learned relative pose regression (RPR) estimate the query pose against a posed reference database, typically sparse and reconstructed offline via SfM~\cite{colmap}. They lead long-term localization benchmarks~\cite{sattler2018benchmark} and are our focus.

Classical pipelines~\cite{hloc} chain local feature extraction~\cite{aliked}, matching~\cite{lightglue}, and robust Perspective-$n$-Point (PnP)~\cite{epnp,ransac}. Recent feed-forward networks replace this pipeline with a single forward pass, predicting relative poses or pointmaps for two-view~\cite{monst3r,reloc3r} or multi-view~\cite{vggt-omega,pi3} sets. 
While these models reduce system complexity, most tolerate aliased reference sets worse than classical pipelines, even when trained on dynamic scenes~\cite{monst3r, vggt-omega, pi3}.   

\subsubsection{Appearance-Based Retrieval}

Classical and learned relative methods retrieve references with appearance-based Visual Place Recognition (VPR)~\cite{netvlad, Izquierdo_CVPR_2024_SALAD}, which is vulnerable to visual aliasing. In repetitive or changed scenes, retrieved references can be confidently wrong, and no relative localizer recovers~\cite{Lowry2016,sattler2018benchmark}. Existing remedies stay in the appearance domain. Hierarchical pipelines cluster the retrieved frames by co-visibility before matching~\cite{hloc}, and sequence-based VPR~\cite{seqslam} enforces temporal consistency of descriptor matches. Gating the appearance search by a pose prior, which we evaluate as VPR+, still ranks by descriptor similarity and inherits its failure modes. We instead remove appearance from the retrieval decision and rank references by pose.

\subsection{3DGS Localization and SLAM}
\label{sec:rw_3dgs_slam}

\subsubsection{Localizing with Rendered References}

Rendering a scene model for localization predates 3DGS: Torii et al.~\cite{torii247} synthesize database views so that query and reference share a viewpoint, MeshLoc extracts features on mesh renderings~\cite{meshloc}, LENS trains a pose regressor on NeRF views placed by the scene geometry~\cite{lens}, and Ress et al.~\cite{ress2025gs} add 3DGS renders at random poses to an SfM reference set. Loc-NeRF~\cite{locnerf}, GS-CPR~\cite{liu2025gscpr}, and GSVisLoc~\cite{gsvisloc} instead work at query time, respectively: per particle of a Monte Carlo filter, to refine a coarse pose from another localizer, or to refine a direct match of encoded 3D Gaussians at seconds per query. Active localization methods such as ActLoc~\cite{actloc} use the scene model to choose the viewing directions expected to localize best, which does not help a passive camera that cannot be steered. None samples its reference set by robot embodiment and density, pre-encodes it for an off-the-shelf localizer, or changes how references are retrieved. Unlike query-time methods, all of our rendering happens offline.

\subsubsection{Pose Priors in SLAM and Localization}
Using a propagated pose to decide what to match against is standard in robotics. Monte Carlo localization weights pose hypotheses propagated by odometry against a map~\cite{mcl}, and ORB-SLAM2's localization mode tracks a saved map by matching to the map points predicted visible from the propagated pose~\cite{orbslam2}. Visual localization pipelines, in contrast, treat every query independently and retrieve by appearance. \methodname brings the prior into the localization pipeline: it gates and ranks references in a dense, metric memory by a propagated pose and re-estimates the pose absolutely with an off-the-shelf localizer, avoiding error accumulation as in odometry.

\begin{figure}[tbp]
  \centering
  \includegraphics[width=\linewidth]{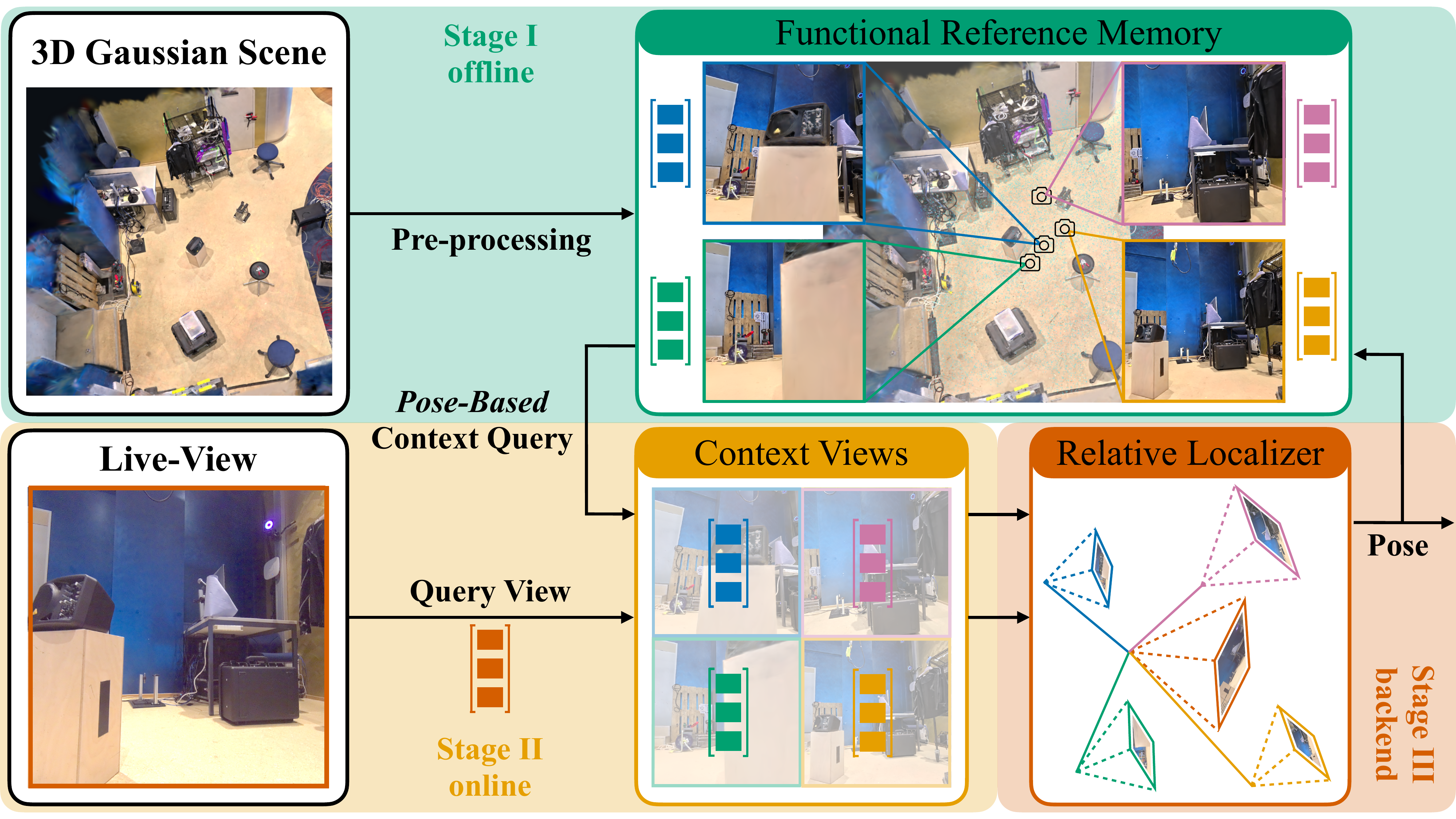}
  \caption{\textbf{\methodname} consists of three stages: an offline scene pre-processing stage that turns the 3D Gaussian Splatting scan into the Functional Reference Memory (\secref{sec:preprocessing}), an online pose-based context retrieval stage (\secref{sec:retrieval}), and an interchangeable backend characterized by a relative localizer (\secref{sec:backends}).
  An appearance-based relocalizer handles initialization and failure recovery.
  }
  \label{fig:system_overview_small}
\end{figure}

\section{System Architecture}
\label{sec:system}

Our proposed system, \methodname~(\figref{fig:system_overview_small}), consists of three stages. The first stage is an offline pre-processing step, where we ingest the 3DGS scan of a scene (\secref{sec:preprocessing}). We sample camera poses \emph{in a controlled manner} within a height band and density informed by the robot's morphology, sensor suite, storage budget, and desired localization accuracy, then render these views and pre-encode them into a format matched to the relative localizer. 
In the second stage, at runtime, we query context encodings from this FRM using a pose prior (\secref{sec:retrieval}) from the previous localization result or, if available, odometry. 
In the third stage, we feed these retrieved references, along with the encoded live query view, into one of several backends (\secref{sec:backends}). Each backend is characterized by a relative localizer, which is used to obtain a global pose.
Our system design deliberately adapts the FRM to the backend, rather than the reverse, enabling relative localizers to run off-the-shelf without per-scene training or fine-tuning.

\subsection{Stage I: 3DGS Scene Pre-Processing}
\label{sec:preprocessing}

Our method utilizes a high-fidelity 3DGS scan and at least a rough collision mesh, both readily obtainable with current 3DGS pipelines and commercial scanners. Our real-world scan comes from a handheld LiDAR-camera scanner that outputs both (\secref{sec:setup}). In simulation, the simulator’s scene model replaces the scan.

\subsubsection{Guided Pose Sampling}

We first extract the navigable space of a scan: from the scene's collision mesh, we rasterize a 2D occupancy grid with cell size $W$. We retain cells with floor support and sufficient headroom $H$, remove cells within a clearance margin $M$ of obstacles via erosion, and discard disconnected islands smaller than $I$, which the robot cannot reach. FRM viewpoints are then sampled to match the robot's specific morphology and sensor suite: positions are sampled uniformly over the navigable area within a sensor-height band $B = [b_{\min}, b_{\max}]$ of thickness $D = b_{\max} - b_{\min}$, with orientation ranges tailored to the platform. The number of FRM poses is strictly density-controlled: $N = \rho \cdot A \cdot D$ for a target density $\rho$ (poses per \si{\cubic\metre}) over navigable area $A$. Density is the primary knob that trades accuracy against storage (\secref{sec:ablations}).

\subsubsection{Rendering and Backend-Specific Encoding}

Every sampled FRM pose is rendered with color and depth using the target camera’s intrinsics, from the 3DGS scan with a real-time Gaussian rasterizer~\cite{gsplat} for real scenes and directly from the simulator for simulated ones,
so simulated references and queries share one renderer. The FRM content is then pre-encoded to exactly match the input requirements of the relative localizer. 
For the classical backend, we extract sparse feature keypoints and descriptors per reference, lift them to metric 3D via the rendered depth, and persist them alongside the camera pose.
For learned backends, we store pre-encoded, posed image latents. Where necessary, we also store depth. In all cases, we additionally store an image descriptor for appearance-based retrieval.
Thus, we offload the heavy encoder runs for multiple reference views to the pre-processing stage, saving inference time. A position k-d tree and an approximate nearest-neighbor index over the image descriptors complete the FRM as its spatial and appearance indices.

FRM size scales linearly with density: $S_{\mathrm{FRM}} = \rho \cdot A \cdot D \cdot (S_{\mathrm{R}} + S_{\mathrm{ID}})$, where $S_{\mathrm{R}}$ and $S_{\mathrm{ID}}$ are the per-reference encoding and image-descriptor sizes.

\begin{figure}[t!]
    \centering
    \begin{subfigure}[b]{0.31\columnwidth}
        \centering
        \includegraphics[width=\linewidth]{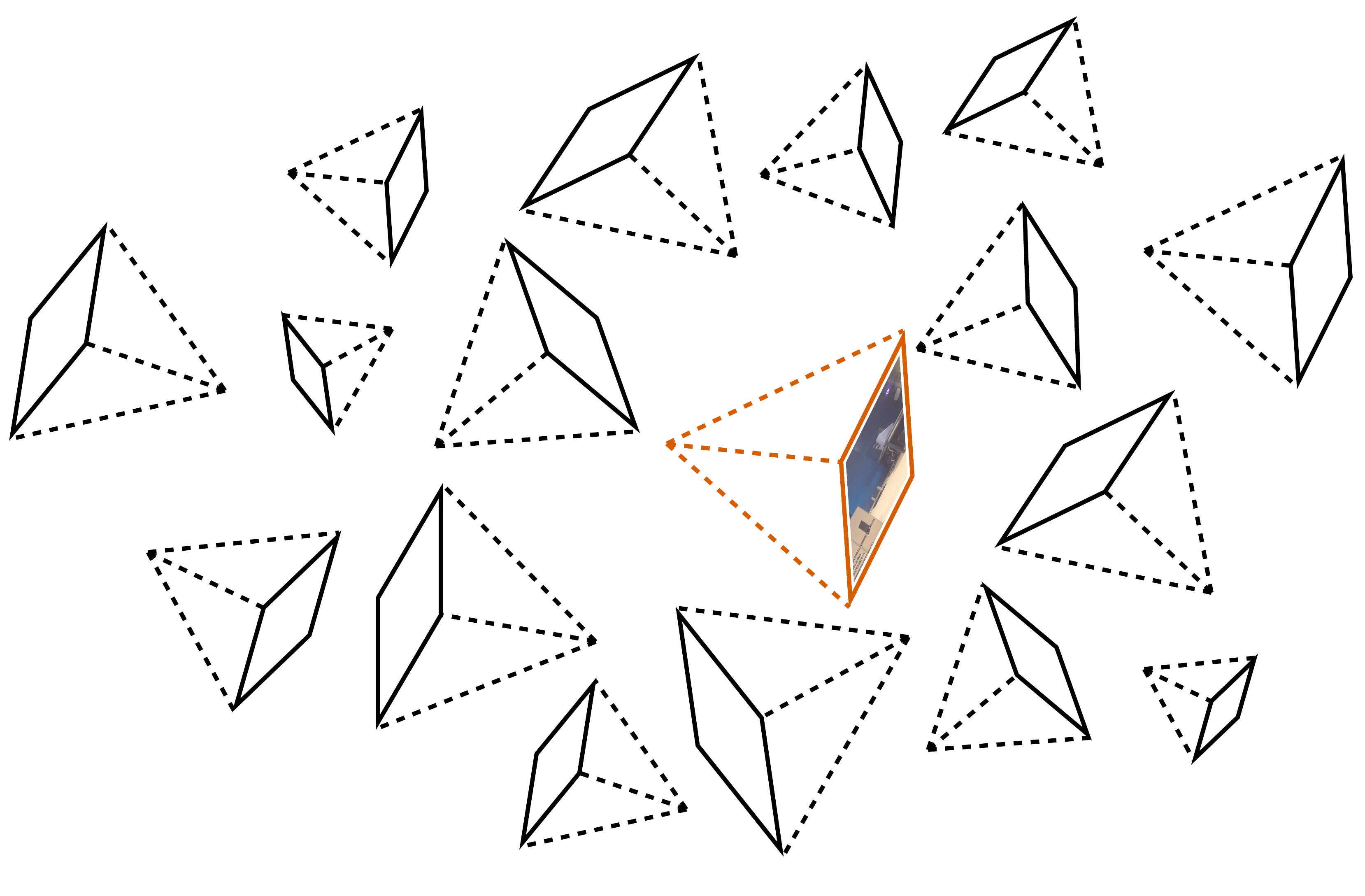}
        \caption{Query image pose}
        \label{fig:query_first}
    \end{subfigure}
    \hfill
    \begin{subfigure}[b]{0.31\columnwidth}
        \centering
        \includegraphics[width=\linewidth]{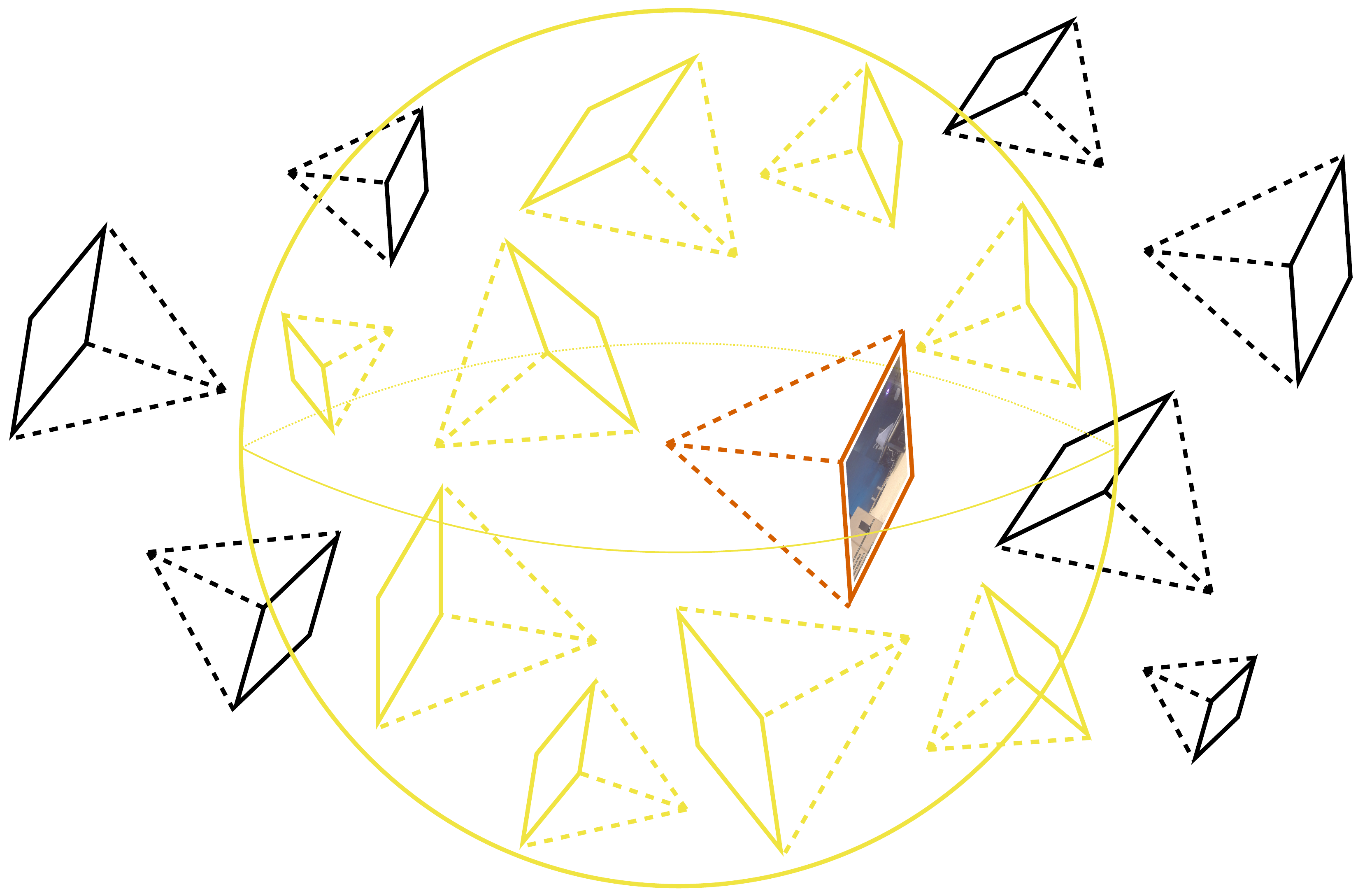}
        \caption{Candidate poses}
        \label{fig:query_second}
    \end{subfigure}
    \hfill
    \begin{subfigure}[b]{0.31\columnwidth}
        \centering
        \includegraphics[width=\linewidth]{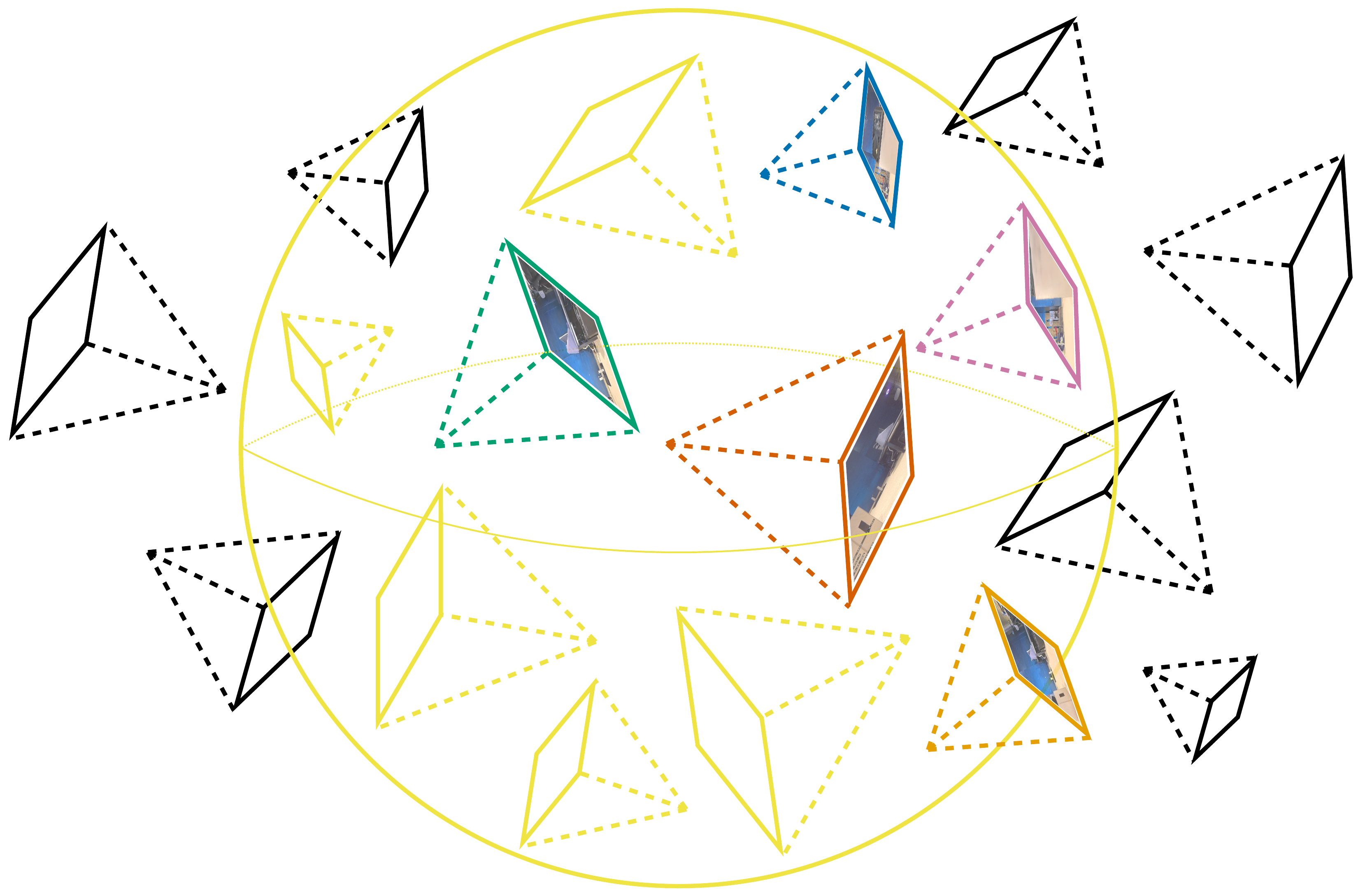}
        \caption{Retrieved references}
        \label{fig:query_third}
    \end{subfigure}

    \caption{
    For each query image (\subref{fig:query_first}), our context retrieval gathers FRM poses within radius $R$ of the current pose guess (\subref{fig:query_second}), keeps the top $P$ by orientation cosine similarity, and selects the $K$ nearest among them (\subref{fig:query_third}). Best viewed in color.
    }
    \label{fig:retrieval}
\end{figure}

\subsection{Stage II: Pose-Based Context Retrieval}
\label{sec:retrieval}

Because robots move continuously, the previous pose estimate can provide a strong prior for the current localization query. However, appearance-based retrieval cannot inherently benefit from such a prior beyond simple gating. In contrast, our retrieval strategy (\figref{fig:retrieval}) queries the FRM's k-d tree for all reference poses within a radius $R$ of the prior position. It then ranks this candidate pool by viewing-direction alignment (cosine similarity between camera forward axes), retains the $P$ best-aligned candidates, and selects the $K$ nearest poses from this subset. Ranking orientation before position ensures the system prioritizes view alignment over spatial proximity.

A query fails only when the backend returns no pose or a non-finite pose, raises an exception, or, for PnP-based backends, fails the minimum-inlier test of \tabref{tab:parameters}. 
We deliberately use no jump or consistency gate, meaning backends that always emit a pose never fall back. 
To initialize, after five consecutive failures, or when fewer than $K$ candidates lie within $R$, the system temporarily falls back to appearance-based retrieval before resuming autoregressive localization.
Notably, our system does not \emph{require} odometry and still beats appearance-based retrieval without it, as we show in \secref{sec:experiments}.

When available, any forward-propagated guess, e.g., from odometry, can replace the previous localization estimate to further improve accuracy and robustness. 
We upper-bound this gain with an oracle prior in \secref{sec:experiments} and make use of odometry in our real-world robot navigation experiment in a highly dynamic scene (\secref{sec:exp_nav}).

\subsection{Stage III: Relative Localization Backends}
\label{sec:backends}

The relative localizer receives the encoded live query image alongside the $K$ references from the FRM and predicts the relative pose. Every method that natively emits scale-normalized output becomes metric solely through the FRM, using known reference poses for alignment and pre-rendered depth for 3D lifting.
We evaluate five different backends:

\emph{Classical multi-stage localizer:} Based on the hloc pipeline~\cite{hloc}, we extract ALIKED~\cite{aliked} features on the live query and match them via LightGlue~\cite{lightglue} directly against the $K$ references' pre-encoded keypoints.
After a per-pair essential-matrix RANSAC filter, we execute a joint PnP~\cite{epnp} plus RANSAC~\cite{ransac} solve over the pooled 2D--3D correspondences, followed by Levenberg--Marquardt (LM) refinement. 

\emph{MonST3R~\cite{monst3r}, Reloc3r~\cite{reloc3r} (two-view feed-forward):} 
For MonST3R, we lift reciprocal pointmap matches between the query and each of the $K$ references to metric 3D using the references' rendered depth, then run PnP with RANSAC on the pooled 2D--3D correspondences using the known query intrinsics.
For Reloc3r, we use its native pipeline: pairwise rotation and translation-direction regression against each of the $K$ references, followed by motion averaging. 

\emph{VGGT-$\Omega$~\cite{vggt-omega}, $\pi^3$~\cite{pi3} (multi-view feed-forward):} We conduct one forward pass over the set of [$K$ references, query]. We align the predicted reference camera centers to their known metric FRM poses (Umeyama alignment with scale~\cite{umeyama}), which maps the predicted query pose into the scene frame. For $\pi^3$, we additionally consume the references' poses, intrinsics, and rendered depth through its native conditioning branch.

All learned backends run with their publicly available weights for the configurations listed in \tabref{tab:parameters} with zero fine-tuning on our scenes.

\section{Experiments}
\label{sec:experiments}

\subsection{Experimental Setup}
\label{sec:setup}

\subsubsection{Evaluation Scenarios}

We compare \methodname with existing baselines across four scenarios: heavy scene changes, increasing human interaction, real-world localization, and in-the-loop with a navigation policy.

\begin{figure}[ht]
    \centering
    \begin{subfigure}[b]{0.49\linewidth}
        \centering
        \includegraphics[width=\linewidth]{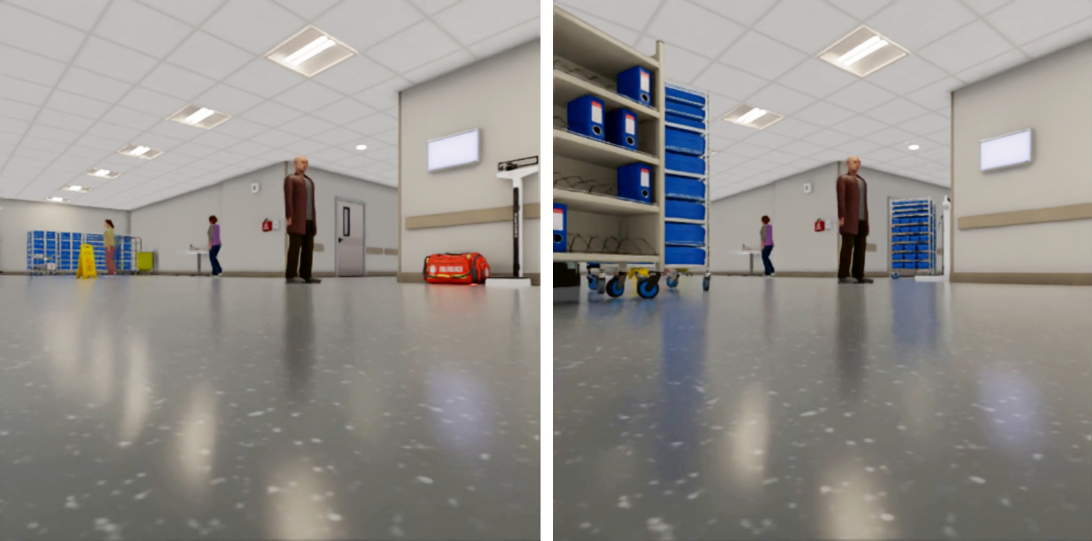}
        \caption{Arena Rosnav hospital}
        \label{fig:arena_bench}
    \end{subfigure}
    \hfill
    \begin{subfigure}[b]{0.49\linewidth}
        \centering
        \includegraphics[width=\linewidth]{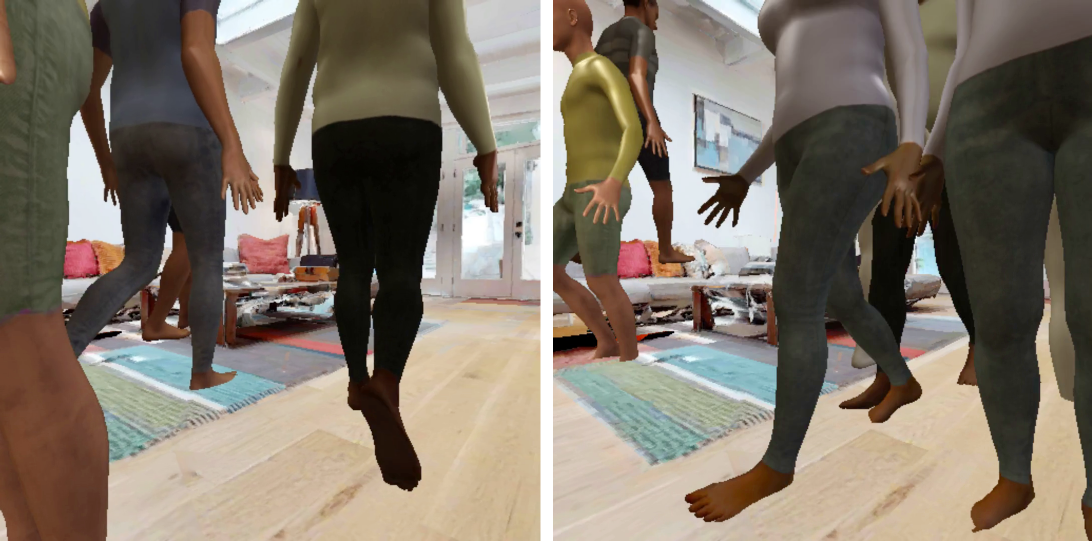}
        \caption{HM3D apartment}
        \label{fig:hm3d_bench}
    \end{subfigure}
    \caption{
    \textbf{Simulated Benchmarks:} Views in the Arena Rosnav hospital and HM3D apartment scenes.
    (\subref{fig:arena_bench})~The same view under scene changes.
    (\subref{fig:hm3d_bench})~The same view with $G=4$ (left) and $G=24$ (right).
    }
    \label{fig:hm3d_arena_benches}
\end{figure}

\emph{Controllable Scene Changes in Arena Rosnav:}
To freely control scene composition, we build two scenes in Arena Rosnav~\cite{ArenaRosnav}: a \SI{384}{\square\metre} hospital wing with 53 movable objects and a \SI{273}{\square\metre} warehouse with 75 objects. In each, a navigation policy follows hand-selected waypoints, shared across both robot platforms (\tabref{tab:platforms}), to produce a physically plausible \SI{97}{\metre}/\SI{98.5}{\metre} robot trajectory (hospital/warehouse), and we sample six human trajectories crossing the robot's field of view (FOV).
Keeping all trajectories fixed, we resample each scene ten times by adding, removing, or repositioning movable objects so they never intersect any path and stay within the robot's FOV along its trajectory.
On average, each resampling moves 24.8 objects in each scene, removes 5.0/7.1 objects (plus 117.8 of 1436 shelf boxes in the warehouse), and adds 2.7/2.9 duplicates.
\figref{fig:arena_bench} shows a view from the Boxer sensor suite in the hospital scene under changes.

\emph{Controllable Human Interactions in HM3D:}
We populate a \SI{37}{\square\metre} region of the real-world HM3D scan 00807~\cite{hm3d} with simulated humans moving as a group, sampling ten trajectory sets for each group size $G \in \{4, 8, 16, 24\}$.
A smooth \SI{24}{\metre} camera trajectory follows the group, fitted through multiple waypoints with a curvature-adaptive speed of up to \SI{0.4}{\metre/\second} and kept within $\pm$\SI{3}{\centi\metre} of the sensor height (\tabref{tab:platforms}).
This controlled following setup measures performance as a function of dynamic view obstruction rather than at a single operating point.
\figref{fig:hm3d_bench} shows a view at the same timestep with different numbers of humans.

\emph{Real-World Localization in a Motion Capture Scene:}
To assess real-world performance and the rendered-to-real gap, we record a dataset combining scene changes and human interaction with our custom bipedal robot on a \SI{14.2}{\square\metre} MoCap stage densely filled with movable objects.
It comprises a two-minute \emph{static} sequence and a three-minute \emph{dynamic} sequence, in which two humans walk through the scene, freely displace objects, and interact with the robot. Both are recorded at \SI{30}{\hertz} and queried offline at \SI{10}{\hertz}, with ground truth from MoCap.
To further exacerbate scene change, we scan the scene only after all recordings, once all objects have been moved, and localize queries from both sequences against the resulting \emph{mismatched} FRM. We scan the scene with a handheld XGRIDS PortalCam and use the same images to form the SfM-posed caches of \secref{sec:exp_mocap}.

\emph{In-The-Loop Navigation in a Real Hallway:}
We run two real-world navigation experiments with \methodname in the loop, using a custom navigation policy in a hallway with repetitive structures where human actors move around the robot and rearrange objects during operation (\figref{fig:teaser}).
We train the policy with hierarchical reinforcement learning following~\cite{xu2025navrl} on top of a standard locomotion policy, and it receives egocentric 2D raycasts and the global pose from our framework.

\subsubsection{Retrieval Conditions}
\label{sec:setup_retrieval}

We evaluate \methodname{} by running each relative localizer (\secref{sec:backends}) under four retrieval conditions over the same FRM. \emph{Ours} is our pose-based retrieval in autoregressive mode, where the previous localization output seeds the next query. \emph{Oracle} seeds our retrieval with the ground-truth query pose and upper-bounds what accurate odometry could contribute. \emph{VPR} is the appearance-based retrieval of standard pipelines. \emph{VPR+} gates VPR to the radius $R$ around its own autoregressive prior, as \emph{Ours} does, but still ranks candidates by descriptor similarity.

In our experiments, the CNN-based NetVLAD~\cite{netvlad} proved more robust to visual aliasing and less volatile overall than the more recent, transformer-based SALAD~\cite{Izquierdo_CVPR_2024_SALAD}. Hence, we use NetVLAD for our VPR baseline and as the reinitialization method for our proposed system.

All four conditions share the identical FRM and, for the classical backend, the same depth-lifted features. \emph{Ours} vs.\ \emph{VPR}/\emph{VPR+} isolates the retrieval rule and \emph{Ours} vs.\ \emph{Oracle} isolates the cost of prior drift. 
\emph{VPR+} specifically separates the benefit of a prior from that of orientation-aware pose ranking.
For the real-world results, we also run backends on caches of real images posed by SfM and compare their performance to that on the FRM.

\begin{table}[tb]
\caption{Evaluated robot platforms and their sensor configurations.
}
\label{tab:platforms}
\centering
\footnotesize
\setlength{\tabcolsep}{4pt}
\begin{tabular}{llcccc}
\toprule
\textbf{Platform} & \textbf{Camera} & \textbf{Height} & \textbf{Pitch} & \textbf{HFOV} & \textbf{Image size} \\
\midrule
\multicolumn{6}{l}{\emph{Simulation (Arena Rosnav)}}\\
TurtleBot      & OAK-D Pro      & \SI{0.179}{\metre} & \ang{0}     & \ang{69}   & $512{\times}512$ \\
Boxer  & RealSense D435 & \SI{0.084}{\metre} & \ang{14.76} & \ang{69.4} & $512{\times}512$ \\
\addlinespace
\multicolumn{6}{l}{\emph{Simulation (HM3D)}}\\
-      & Pinhole      & \SI{0.45}{\metre} & \ang{0}     & \ang{90}   & $512{\times}512$ \\
\addlinespace
\multicolumn{6}{l}{\emph{Real-world (MoCap and Navigation setups, respectively)}}\\
Bipedal  & IMX900         & \SI{0.40}{\metre}  & \ang{0}     & \ang{90}   & $512{\times}384$ \\
Bipedal  & IMX335         & \SI{0.42}{\metre}   & \ang{0}     & \ang{115}  & $640{\times}480$ \\
\bottomrule
\end{tabular}
\end{table}

\begin{table}[tb]
\caption{\methodname{} parameters. Values are shared across all experiments
unless ablated in \secref{sec:ablations}.}
\label{tab:parameters}
\centering
\footnotesize
\setlength{\tabcolsep}{4pt}
\begin{tabular}{llc}
\toprule
& \textbf{Parameter} & \textbf{Value} \\
\midrule
\multirow{8}{*}{\rotatebox{90}{\emph{Functional Reference Memory}}}
 & Occupancy grid cell $W$        & \SI{0.05}{\metre} \\
  & Obstacle clearance $M$        & \SI{0.05}{\metre} \\
 & Required headroom $H$          & \SI{1.8}{\metre} \\
 & Min.\ island area $I$          & \SI{0.5}{\square\metre} \\
 & Height Band $B$ (Arena, TurtleBot)       & [\SI{5}{\centi\metre}, \SI{32.9}{\centi\metre}] \\
 & Height Band $B$ (Arena, Boxer)           & [\SI{5}{\centi\metre}, \SI{23.4}{\centi\metre}] \\
 & Height Band $B$ (HM3D \& real-world)     & [\SI{20}{\centi\metre}, \SI{1.2}{\metre}] \\
 & Pitch / roll / yaw$^\dagger$ (Arena)     & $\pm\ang{4}$ / $\pm\ang{0}$ / $\pm\ang{180}$ \\
 & Pitch / roll / yaw$^\dagger$ (HM3D \& real-world)     & $\pm\ang{30}$ / $\pm\ang{5}$ / $\pm\ang{180}$ \\
 & Density $\rho$                 & \SI{1000}{\per\cubic\metre} \\
 & Render size                    & $512^2$ \\
 & Keypoints per reference        & $\le 256$ (thr.\ 0.2) \\
\midrule
\multirow{4}{*}{\rotatebox{90}{\emph{Retrieval}}}
 & Query radius $R$               & \SI{1.5}{\metre} \\
 & Orientation pool $P$           & 64 \\
 & Context size $K$               & classical: 4 / learned: 8 \\
 & Relocalization trigger         & 5 consecutive failures \\
\midrule
\multirow{3}{*}{\rotatebox{90}{\emph{Solver}}}
 & PnP solver                     & EPnP + LM \\
 & Reprojection threshold         & \SI{4}{px} \\
 & Minimum inliers per reference  & 15 \\
\midrule
\multirow{4}{*}{\rotatebox{90}{\emph{Backends}}}
 & MonST3R                        & 512-DPT, conf.\ 3.0 \\
 & Reloc3r                        & 512, mixed precision \\
 & VGGT-$\Omega$                  & 1B/512 \\
 & $\pi^3$                        & $504^2$ \\
\bottomrule
\addlinespace[2pt]
\multicolumn{3}{@{}l}{$^\dagger$Pitch, roll, and yaw sample ranges are relative to the sensor mount.} \
\end{tabular}
\end{table}

\subsubsection{Platforms and Parameters}
\label{sec:setup_platforms}
\tabref{tab:platforms} lists the five evaluated sensor configurations, three simulated and two on the real robot. \tabref{tab:parameters} lists the method parameters, which are shared across all experiments unless ablated in \secref{sec:ablations}. We define FRM height bands per platform to encompass the mounted camera, so one scan can serve different robot morphologies with different mounts.

\subsubsection{Metrics} 
We report recall@\SI{5}{\centi\metre}/\ang{5} and position root mean squared error (RMSE). Recall is computed over \emph{all} ground-truth frames. This way, an unlocalized frame counts as a miss and coverage is folded into the metric. RMSE is calculated over \emph{valid} frames and exposes outlier proneness~\cite{sattler2018benchmark}. Recall can improve at the expense of RMSE and vice versa; when a method solves more frames at the cost of more outliers, this does not directly indicate greater robustness. 

\subsection{Simulated Scene Changes}
\label{sec:exp_arena}

\begin{figure}[tbp]
  \centering
  \includegraphics[width=\linewidth]{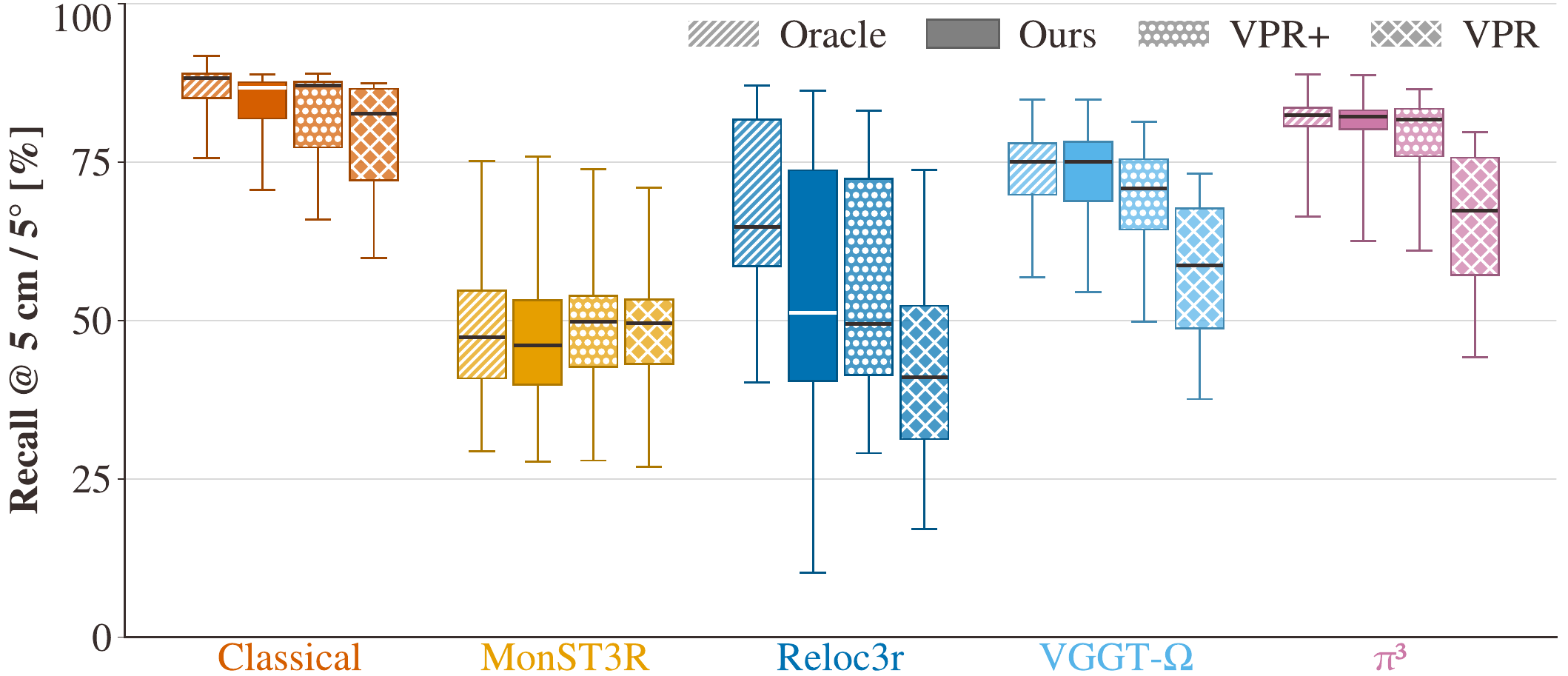}
    \captionof{figure}{
    \textbf{Simulated Dynamic Scenes:} Localization performance on two Arena Rosnav scenes with two sensor setups and ten object rearrangements each. Our context retrieval improves most localization backends except for MonST3R, while oracle pose priors mainly raise the recall floor and reduce variance further.
    }
    \label{fig:arena_rosnav_dynamic_scenes_pooled}
\end{figure}

This experiment tests whether pose-based retrieval survives object rearrangement that defeats appearance retrieval.
\figref{fig:arena_rosnav_dynamic_scenes_pooled} shows the localization performance of all evaluated backends with varying retrieval methods on the Arena Rosnav scenes. 

For four of five backends, our retrieval outperforms VPR and VPR+ by raising the median recall and/or the recall floor (MonST3R is on par).
The prior accounts for most of the gain, and orientation-aware ranking adds further improvement.
Furthermore, a better pose prior, as assumed in oracle retrieval, mainly raises the recall floor, leading to slightly higher and tighter overall performance.

Notably, most learned backends degrade with plain VPR retrieval.
On the Arena scenes, VPR-retrieved references lie a median \SI{0.78}{\metre} from the ground-truth query pose and \SI{1.01}{\metre} apart from each other, versus \SIrange{0.28}{0.37}{\metre} and \SIrange{0.36}{0.49}{\metre} for our pose-based retrieval, indicating visual aliasing.
We hypothesize that Reloc3r, VGGT-$\Omega$, and $\pi^3$ are sensitive to a mix of correct \emph{and} incorrect references.
The classical backend and MonST3R show less performance loss with VPR, likely because their confidence estimates let them discount bad references.
VPR+ removes most of this loss, but most backends benefit from the consistent reference sets our retrieval provides.

\subsection{Simulated Human Activity}
\label{sec:exp_hm3d}

\begin{figure}[tbp]
  \centering
  \includegraphics[width=\linewidth]{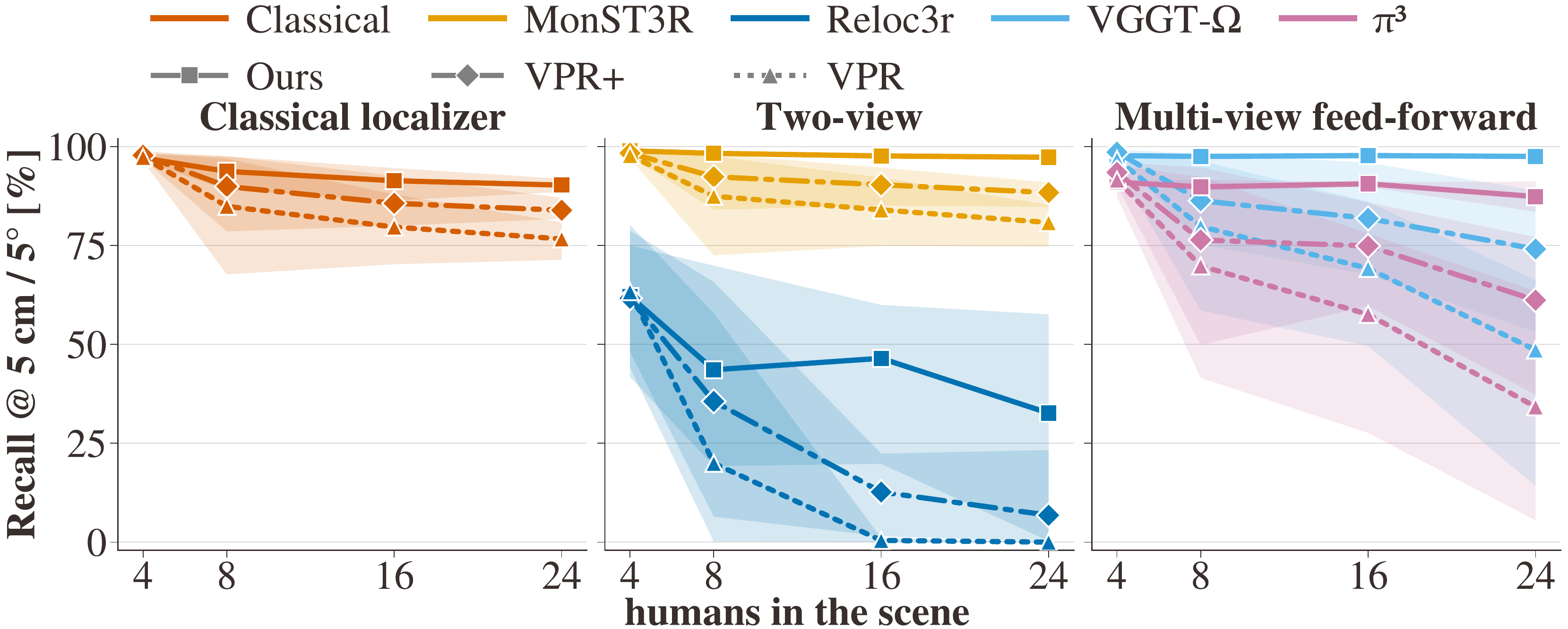}
    \captionof{figure}{
    \textbf{Simulated Human Interactions:} Mean localization performance with P10/P90 bands versus the number of simulated humans in the HM3D human-following scenario. For all localizers, VPR \& VPR+ retrieval degrades markedly as we add humans and varies more across seeds, while ours is largely unaffected, except for Reloc3r.
    }
    \label{fig:hm3d_humans_follow}
\end{figure}

This experiment tests how retrieval is affected by the number of dynamic distractors.
\figref{fig:hm3d_humans_follow} shows the performance trends of the evaluated backends on the HM3D scene.

As the group size $G$ grows from 4 to 24, four of the five backends with our retrieval lose at most seven recall points, and Reloc3r drops from 62.1\,\% to 32.6\,\%, whereas every backend with VPR+ or VPR loses more, down to 6.8\,\% and 0\,\% for Reloc3r.

Backends using our retrieval also show a smaller P10/P90 spread across distractor samplings. This indicates that appearance retrieval is not only less accurate but also less predictable under visual distractors.

\subsection{Real-World Localization Benchmark}
\label{sec:exp_mocap}

\definecolor{posgreen}{HTML}{009e73}
\definecolor{negred}{HTML}{d55e00}
\newcommand{\pos}[1]{\textsuperscript{\textcolor{posgreen}{#1}}}
\newcommand{\negv}[1]{\textsuperscript{\textcolor{negred}{#1}}}

This experiment tests the full system on real queries and compares rendered against real references.
We summarize the recall and RMSE localization results for the different backends on our MoCap scene in \tabref{tab:robot_metrics_moved}.

Relative to VPR, our retrieval lowers position RMSE in nine of ten backend/split combinations by 15--60\,\%, and raises recall for the classical, MonST3R, and Reloc3r backends by 2--27\,\%. For VGGT-$\Omega$ and $\pi^3$, recall changes by $-6.6$ to $+2.2$\,\%, and $\pi^3$ in the \emph{dynamic} split is the one case where RMSE rises. Lower RMSE at similar recall means fewer outliers. VPR+ recovers only part of the RMSE gain.

We attribute the smaller gaps relative to simulation (Figs.~\ref{fig:arena_rosnav_dynamic_scenes_pooled} and~\ref{fig:hm3d_humans_follow}) to the small \SI{14.2}{\square\metre} scene, which offers little aliasing, and to having only two human distractors.

The lower block runs the classical localizer on two caches of real images of the same moved scene the FRM was built on, now fully mirroring hloc~\cite{hloc}. We pose the PortalCam images from which the 3DGS was built with COLMAP~\cite{colmap} (exhaustive matching, rig extrinsics fixed, intrinsics initialized from vendor values and refined), align them to the scanner’s LiDAR-inertial odometry trajectory by one Sim(3) Umeyama alignment~\cite{umeyama}, and triangulate keypoints with no rendered views or depth. \emph{Single cam} keeps the left pinhole camera's 323 images ($\sim$\SI{18}{m^{-3}} over the \SI{17.7}{\metre^3} capture volume), \emph{four cam rig} all 1160 ($\sim$\SI{66}{m^{-3}} over the same volume).

\begin{figure}[!t]
  \centering
  \includegraphics[width=\linewidth]{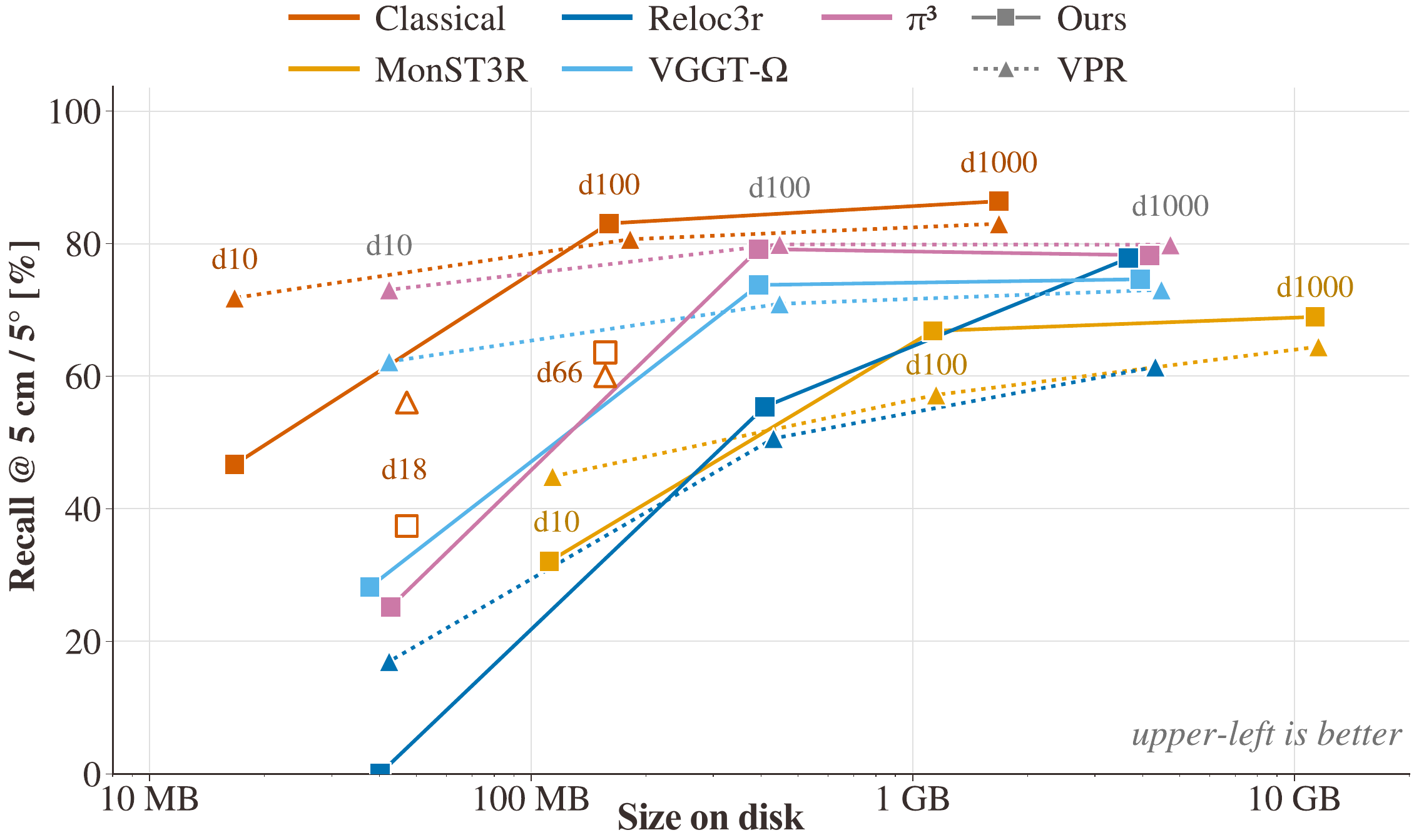}
    \captionof{figure}{
    \textbf{Real-World On-Robot Storage:} FRM density--storage trade-off per backend on the \emph{dynamic} MoCap split. Denser references help all localizers under both retrieval methods, but our pose-based retrieval benefits from a sufficiently dense FRM ($\geq$\SI{100}{\metre^{-3}}).
    Hollow markers: the classical backend on the SfM-posed real-image caches.
    }
    \label{fig:cache_density_ablation}
\end{figure}

In simulation, references and queries come from the same renderer.
The MoCap experiment is the only one that pits real queries against both rendered and real references, shown in the upper and lower block of \tabref{tab:robot_metrics_moved}.
With the same pipeline, real queries localize better against our FRM (0.864 recall, \SI{8.2}{\centi\metre} RMSE, dynamic) than against SfM-posed real images of the same scene (at best 0.636 and \SI{16.6}{\centi\metre}). Additionally, in \figref{fig:cache_density_ablation} an FRM of only \SI{10}{m^{-3}} queried by the classical backend with VPR already exceeds the performance with the  \SI{66}{m^{-3}} rig cache.
The performance cost of the appearance gap between rendered references and real query images is thus outweighed by the gains from controlled, embodiment-specific viewpoint coverage. Acquisition effort is identical, since the FRM and the SfM caches come from the same scan.

We also note that while our system improves most backends, the classical pipeline combines the highest recall in both splits (above \SI{80}{\percent}) with an RMSE below \SI{10}{\centi\metre}, and would be our recommendation for real-world deployment.

\begin{table}[tb]
\caption{
\textbf{Localization evaluation on real-world MoCap data:} Recall@5\,cm/5$^\circ$ and position RMSE per backend. Superscripts: relative change w.r.t. the VPR baseline (\textcolor{posgreen}{green} better, \textcolor{negred}{orange} worse). Best per column within each block in \textbf{bold}. 
Our retrieval improves nearly all localizers in RMSE, recall, or both. On the FRM, the classical backend beats the same pipeline on SfM-posed real-image caches in every column.}

\label{tab:robot_metrics_moved}
\centering
\footnotesize
\setlength{\tabcolsep}{6.pt}
\begin{tabular}{ll cccc}
\toprule
& & \multicolumn{2}{c}{\textbf{Static}} & \multicolumn{2}{c}{\textbf{Dynamic}} \\
\cmidrule(lr){3-4} \cmidrule(lr){5-6}
\textbf{Method} & \textbf{Retr.} & R@5 & RMSE [cm] & R@5 & RMSE [cm] \\
\midrule
\multicolumn{6}{l}{\textit{FRM (ours, rendered images, 3DGS based, \SI{1000}{m^{-3}})}} \\[1pt]
Classical
  & Ours & \textbf{0.889}\rlap{\pos{$\uparrow$1.8\%}}     & \textbf{6.3}\rlap{\pos{$\downarrow$34.4\%}}    & \textbf{0.864}\rlap{\pos{$\uparrow$4.1\%}}              & 8.2\rlap{\pos{$\downarrow$54.2\%}}             \\
  & VPR+ & 0.878\rlap{\pos{$\uparrow$0.6\%}}              & 6.7\rlap{\pos{$\downarrow$30.2\%}}             & 0.840\rlap{\pos{$\uparrow$1.2\%}}              & 9.7\rlap{\pos{$\downarrow$45.8\%}}             \\
  & VPR & 0.873                                          & 9.6                                            & 0.830                                          & 17.9                                           \\
\hdashline\noalign{\vskip 2.5pt}
MonST3R
  & Ours & 0.729\rlap{\pos{$\uparrow$7.2\%}}              & 7.3\rlap{\pos{$\downarrow$54.1\%}}             & 0.689\rlap{\pos{$\uparrow$6.8\%}}              & \textbf{7.4}\rlap{\pos{$\downarrow$43.9\%}}    \\
  & VPR+ & 0.681\rlap{\pos{$\uparrow$0.1\%}}              & 14.5\rlap{\pos{$\downarrow$8.8\%}}             & 0.664\rlap{\pos{$\uparrow$2.9\%}}              & 16.1\rlap{\negv{$\uparrow$22.0\%}}             \\
  & VPR & 0.680                                          & 15.9                                           & 0.645                                          & 13.2                                           \\
\hdashline\noalign{\vskip 2.5pt}
Reloc3r
  & Ours & 0.826\rlap{\pos{$\uparrow$15.5\%}}             & 26.9\rlap{\pos{$\downarrow$51.2\%}}            & 0.778\rlap{\pos{$\uparrow$26.7\%}}             & 16.3\rlap{\pos{$\downarrow$59.7\%}}            \\
  & VPR+ & 0.727\rlap{\pos{$\uparrow$1.7\%}}              & 52.1\rlap{\pos{$\downarrow$5.4\%}}             & 0.652\rlap{\pos{$\uparrow$6.2\%}}              & 33.1\rlap{\pos{$\downarrow$18.1\%}}            \\
  & VPR & 0.715                                          & 55.1                                           & 0.614                                          & 40.4                                           \\
\hdashline\noalign{\vskip 2.5pt}
VGGT-$\Omega$
  & Ours & 0.690\rlap{\negv{$\downarrow$6.6\%}}           & 41.6\rlap{\pos{$\downarrow$20.6\%}}            & 0.746\rlap{\pos{$\uparrow$2.2\%}}              & 43.4\rlap{\pos{$\downarrow$15.1\%}}            \\
  & VPR+ & 0.746\rlap{\pos{$\uparrow$0.9\%}}              & 57.0\rlap{\negv{$\uparrow$8.8\%}}              & 0.752\rlap{\pos{$\uparrow$3.0\%}}              & 49.9\rlap{\pos{$\downarrow$2.3\%}}             \\
  & VPR & 0.739                                          & 52.4                                           & 0.730                                          & 51.1                                           \\
\hdashline\noalign{\vskip 2.5pt}
$\pi^3$
  & Ours & 0.807\rlap{\negv{$\downarrow$1.7\%}}           & 31.9\rlap{\pos{$\downarrow$21.8\%}}            & 0.782\rlap{\negv{$\downarrow$2.0\%}}           & 34.8\rlap{\negv{$\uparrow$17.2\%}}             \\
  & VPR+ & 0.827\rlap{\pos{$\uparrow$0.7\%}}              & 34.7\rlap{\pos{$\downarrow$15.0\%}}            & 0.804\rlap{\pos{$\uparrow$0.8\%}}              & 33.4\rlap{\negv{$\uparrow$12.5\%}}             \\
  & VPR & 0.821                                          & 40.8                                           & 0.798                                          & 29.7                                           \\

\midrule
\multicolumn{6}{l}{\textit{Sparse reference cache (real images, posed with COLMAP~\cite{colmap})}} \\[1pt]
Classical
  & Ours & 0.396\rlap{\negv{$\downarrow$39.6\%}}          & 32.7\rlap{\negv{$\uparrow$7.9\%}}              & 0.374\rlap{\negv{$\downarrow$33.2\%}}          & 43.3\rlap{\pos{$\downarrow$35.9\%}}            \\
{\scriptsize single cam}  & VPR+ & 0.664\rlap{\pos{$\uparrow$1.2\%}}              & 21.1\rlap{\pos{$\downarrow$30.4\%}}            & 0.598\rlap{\pos{$\uparrow$6.8\%}}              & \textbf{16.6}\rlap{\pos{$\downarrow$75.4\%}}   \\
\SI{18}{m^{-3}}  & VPR & 0.656                                          & 30.3                                           & 0.560                                          & 67.5                                           \\
\hdashline\noalign{\vskip 2.5pt}
Classical
  & Ours & 0.649\rlap{\negv{$\downarrow$6.9\%}}           & \textbf{13.9}\rlap{\pos{$\downarrow$32.2\%}}   & \textbf{0.636}\rlap{\pos{$\uparrow$6.0\%}}     & 31.1\rlap{\pos{$\downarrow$28.7\%}}            \\
{\scriptsize four cam rig}  & VPR+ & \textbf{0.708}\rlap{\pos{$\uparrow$1.6\%}}     & 15.5\rlap{\pos{$\downarrow$24.4\%}}            & 0.615\rlap{\pos{$\uparrow$2.5\%}}              & 21.1\rlap{\pos{$\downarrow$51.6\%}}            \\
\SI{66}{m^{-3}}  & VPR & 0.697                                          & 20.5                                           & 0.600                                          & 43.6                                           \\
\bottomrule

\end{tabular}
\end{table}

\subsection{Robust Navigation in an Actively Changing Scene}
\label{sec:exp_nav}

This experiment is a qualitative end-to-end demonstration of the full system supporting a real navigation task in a busy scene.
Given its real-world performance (\secref{sec:exp_mocap}), we deploy only the classical backend for hallway navigation, running onboard at \SI{5}{\hertz} with odometry-propagated pose priors (closer to the \emph{Oracle} condition than \emph{Ours}). Across two experiments with four and three goals spaced \SI{30}{\metre} apart, the robot \emph{reaches every goal} despite increasing human interference and objects being added and moved (\figref{fig:teaser}).

We refer the reader to the accompanying video for a more in-depth qualitative analysis and scene inspection of the navigation experiment. There, we also show visual aliasing examples of the appearance-based context retrieval methods.

\begin{figure}[!t]
  \centering
  \includegraphics[width=\linewidth]{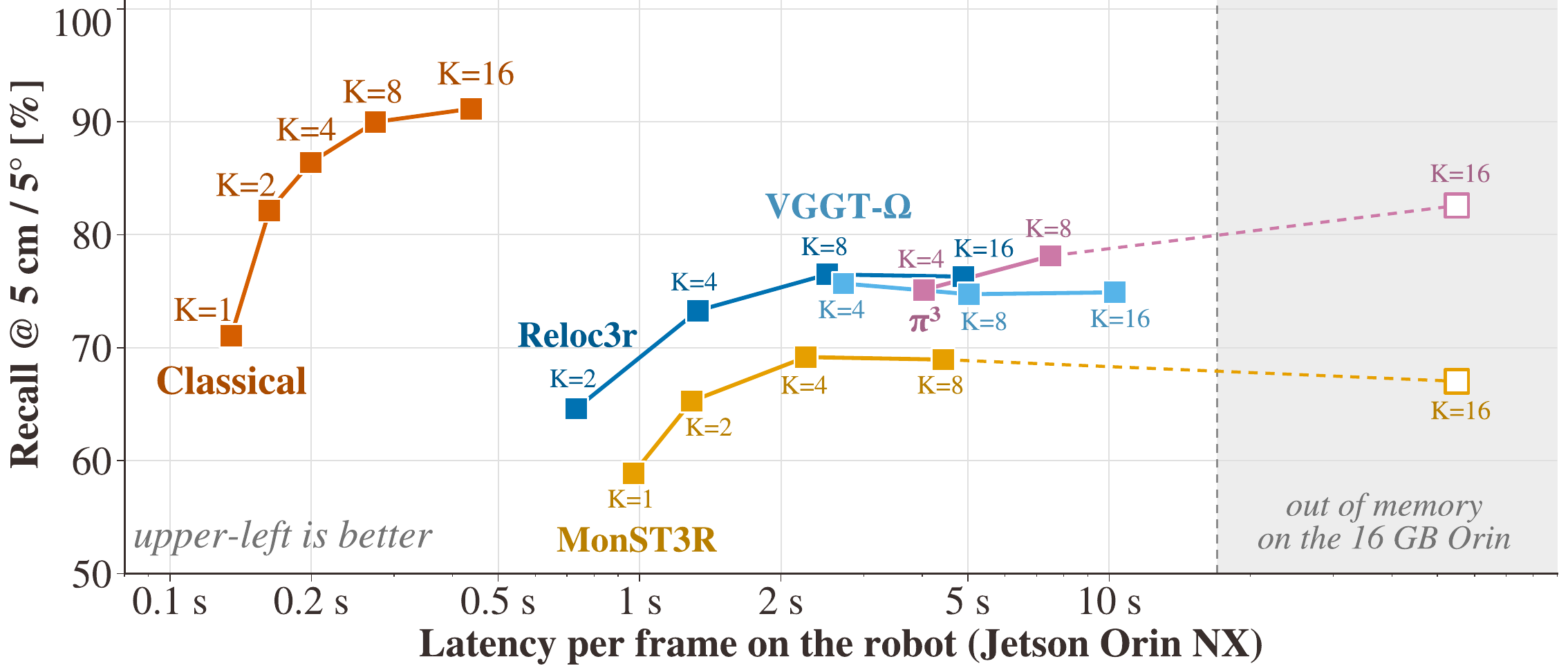}
    \captionof{figure}{
    \textbf{Real-World On-Robot Inference:} Inference time--performance trade-off per backend for different context sizes $K$ with our retrieval on the \emph{dynamic} MoCap split. Only the classical backend with $K=4$ reaches the targeted \SI{5}{\hertz} onboard, while the best trade-off for most localizers lies between $K=4$ and $K=8$.
    }
    \label{fig:context_count_ablation}
\end{figure}

\subsection{Ablations}
\label{sec:ablations}

\subsubsection{FRM Density Ablation}
\label{sec:ablation_density}

We ablate FRM reference density on the \emph{dynamic} real-world split, measuring performance with all retrieval methods along with storage footprint.
As shown in \figref{fig:cache_density_ablation}, denser FRMs help all localizers regardless of retrieval method. On this split, pose-based retrieval needs at least \SI{100}{\metre^{-3}} to beat VPR for all but $\pi^3$, since sparser FRMs lack well-aligned references within the query radius, even when tailored to the robot’s morphology.
Yet already at \SI{10}{\metre^{-3}}, the classical backend with VPR achieves higher recall on our FRM than on either SfM-posed real-image cache (hollow markers).

A light FRM is thus already preferable to SfM-posed caches, and appearance-based retrieval remains advantageous when only a sparse reference set is available. When a dense FRM is affordable, pose-based retrieval is the better choice, particularly in challenging scenes.

Storage scales linearly with density. With the conservative \SI{1}{\metre} height band of \tabref{tab:parameters}, which lets one FRM serve several camera heights, an FRM at \SI{1000}{\metre^{-3}} requires \SI{120}{\mega\byte} per \si{\square\metre} of floor area for the classical backend (\SI{1.7}{\giga\byte} for our \SI{14.2}{\square\metre} scene), of which \SI{16}{\mega\byte} per \si{\metre^2} are VPR descriptors.
Narrowing the band to account for a single morphology and camera height reduces this proportionally. For demanding, changing scenes we recommend a density of \SI{100}{}--\SI{1000}{\metre^{-3}} at a height band of \SI{10}{\centi\metre}, about \SI{1.2}{}--\SI{12}{\mega\byte} per \si{\square\metre}. Note also that our FRM does not need to be sampled uniformly across a large space, so storage can be budgeted per region.

\subsubsection{Reference Number and Runtime Ablation}
\label{sec:ablation_runtime}

\figref{fig:context_count_ablation} displays end-to-end inference time on the robot's onboard computer (Jetson Orin NX, \SI{16}{\giga\byte}) for $K \in \{1, 2, 4, 8, 16\}$, alongside performance on the \emph{dynamic} MoCap split. 
Nearly all backends show a predictable time--performance trade-off, with the best balance for most between $K=4$ and $K=8$, while multi-view feed-forward models typically fail for $K\leq2$; MonST3R and $\pi^3$ run out of memory at $K = 16$ (shaded region in \figref{fig:context_count_ablation}). Only the classical backend with $K=4$ reaches the \SI{5}{\hertz} our navigation policy uses.

Our retrieval takes $\sim$\SI{4}{\milli\second} versus $\sim$\SI{40}{\milli\second} for VPR at a density of \SI{1000}{\metre^{-3}}, though learned forward passes still dominate inference time. Pre-encoding the rendered reference views into our FRM also avoids $K$ ALIKED~\cite{aliked} ($\sim$\SI{70}{\milli\second} each) or DINO~\cite{simeoniDINOv32025} ($\sim$\SI{200}{\milli\second} each) forward passes at runtime, although they can be batched for efficiency.

\section{Conclusion}

We presented \methodname, which turns a metric 3D Gaussian Splatting scan into a dense \emph{Functional Reference Memory} for localization: embodiment-aware reference views, rendered offline and pre-encoded for off-the-shelf localizers. On real data, this memory outperforms SfM-posed real-image caches of the same scene. Pose-based retrieval lowers position RMSE by \SI{15}{}--\SI{60}{\percent} over appearance retrieval in nine of ten real-world backend/split combinations.
For most backends, it handles scene change and human activity better than even prior-gated appearance retrieval.
We also demonstrate the system in-the-loop for navigation in a highly dynamic hallway.

Because the FRM is rendered rather than captured, it can be edited and extended after the scan. \methodname{} already works with scans of furnished rooms, and segmenting movable objects from the scan and rendering references from their occupied space would keep coverage complete after rearrangement. Online FRM updates could track long-term change, and compression would scale FRMs to entire buildings.
Characterizing the rendered-to-real gap under increasing image degradation would inform camera and scan requirements.
Finally, the formulation is not specific to indoor scenes, making outdoor evaluation under changing illumination and seasons a natural next step.




\section*{ACKNOWLEDGMENT}

Anthropic's Claude served as an editorial tool for language and consistency. All research content is the authors' own.


\bibliographystyle{IEEEtran}
\bibliography{root}

\end{document}